\pdfoutput=1

\documentclass[11pt]{article}

\usepackage[final]{latex/acl}

\usepackage{times}
\usepackage{latexsym}

\usepackage[T1]{fontenc}

\usepackage[utf8]{inputenc}

\usepackage{microtype}

\usepackage{inconsolata}

\usepackage{graphicx}

\usepackage{booktabs}
\usepackage{multirow}
\usepackage{amsmath}
\usepackage{amssymb}
\usepackage{amsfonts}
\usepackage{float}
\newcommand{\sbt}{\,\begin{picture}(-1,1)(-1,-3)\circle*{3}\end{picture}\ }

\title{$\infty$-MoE: Generalizing Mixture of Experts to Infinite Experts}

\author{
 \textbf{Shota Takashiro},
 \textbf{Takeshi Kojima},
\\
 \textbf{Shohei Taniguchi},
 \textbf{Yusuke Iwasawa},
 \textbf{Yutaka Matsuo}
\\
 \textsuperscript{}{The University of Tokyo, Hongo 7-3-1, Bunkyo-ku, Tokyo, 113-8656 Japan}
\\
 \small{
   \{\href{mailto:takashiro@weblab.t.u-tokyo.ac.jp}{takashiro},
    \href{mailto:taniguchi@weblab.t.u-tokyo.ac.jp}{taniguchi},
    \href{mailto:t.kojima@weblab.t.u-tokyo.ac.jp}{t.kojima},
    \href{mailto:iwasawa@weblab.t.u-tokyo.ac.jp}{iwasawa},
    \href{mailto:matsuo@weblab.t.u-tokyo.ac.jp}{matsuo}\}\href{mailto:takashiro@weblab.t.u-tokyo.ac.jp}{@weblab.t.u-tokyo.ac.jp}
 }
}

\begin{document}
\maketitle

\begin{abstract}
The Mixture of Experts (MoE) selects a few feed-forward networks (FFNs) per token, achieving an effective trade-off between computational cost and performance. 
In conventional MoE, each expert is treated as entirely independent, and experts are combined in a discrete space. As a result, when the number of experts increases, it becomes difficult to train each expert effectively. 
To stabilize training while increasing the number of experts, we propose $\infty$-MoE that selects a portion of the parameters of large FFNs based on continuous values sampled for each token. 
By considering experts in a continuous space, this approach allows for an infinite number of experts while maintaining computational efficiency. Experiments show that a GPT-2 Small-based $\infty$-MoE model, with 129M active and 186M total parameters, achieves comparable performance to a dense GPT-2 Medium with 350M parameters. Adjusting the number of sampled experts at inference time allows for a flexible trade-off between accuracy and speed, with an improvement of up to 2.5\% in accuracy over conventional MoE.
\end{abstract}

\begin{figure*}[t!]
    \centering
    \includegraphics[width=1.0\linewidth]{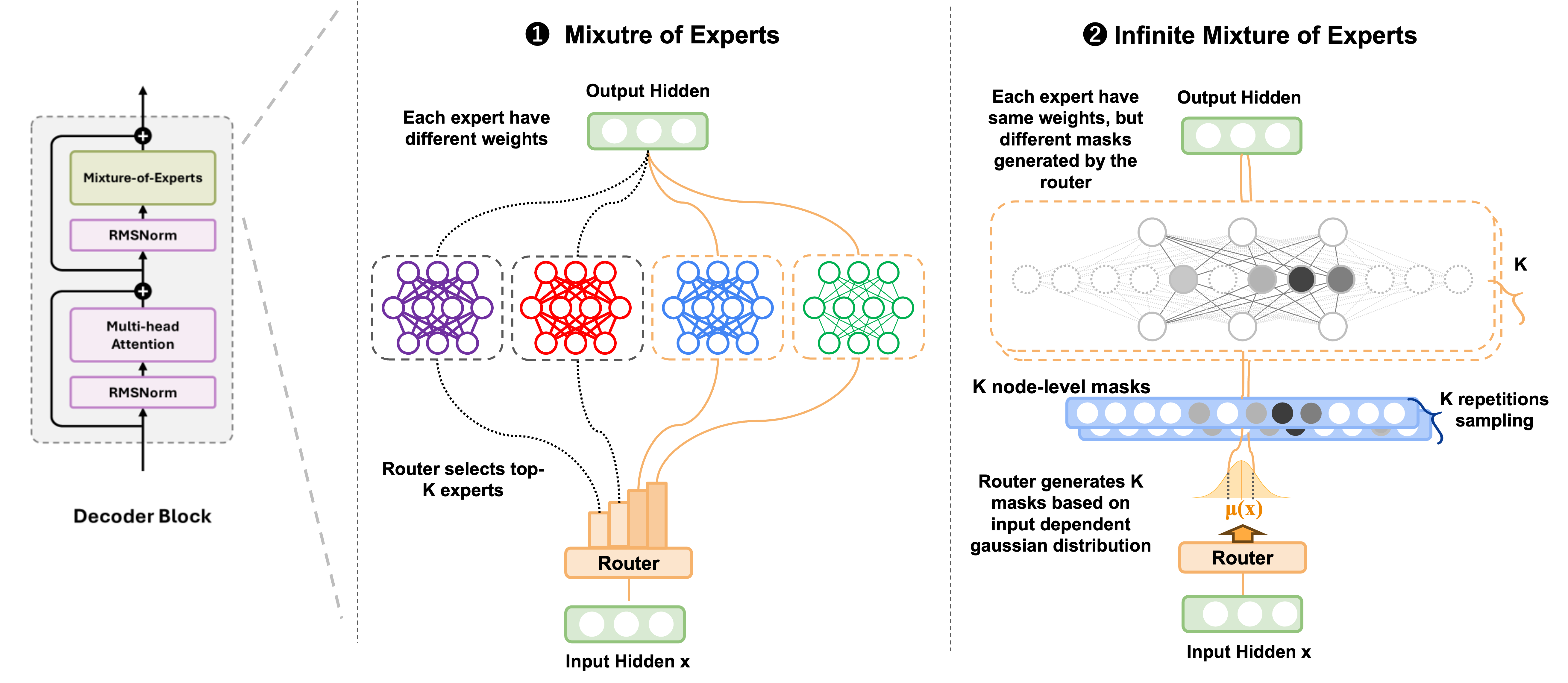}
    \caption{Overview of the proposed Infinite Mixture of Experts ($\infty$-MoE). The router outputs a continuous distribution over the expert space, and each sample selects a unique expert.}
    \label{fig:IMoE}
\end{figure*}

\section{Introduction}

Large language models (LLMs) have recently achieved remarkable performance across a broad range of natural-language processing (NLP) tasks, such as machine translation, question answering, and code generation~\cite{chen2021evaluating, liu2021pre}. These advances are primarily driven by scaling up model parameters, training data, and compute resources~\cite{DBLP:journals/corr/abs-2001-08361}. However, simply increasing model size leads to substantial computational and memory overheads, motivating research into more efficient strategies for scaling.

Mixture of Experts (MoE)~\cite{shazeer2017outrageouslylargeneuralnetworks} stands out for its ability to expand parameter count while maintaining relatively low per-token compute costs. By routing each input to a subset of specialized experts , MoE-based architectures can efficiently store large amounts of knowledge sparsely~\cite{dai2024deepseekmoeultimateexpertspecialization, jiang2024mixtralexperts}. Recent large-scale models such as DeepSeek~\cite{dai2024deepseekmoeultimateexpertspecialization}, Mistral~\cite{jiang2024mixtralexperts}, and Phi~\cite{abdin2024phi3technicalreporthighly} have successfully adopted MoE designs, demonstrating that sparse routing can significantly improve performance without incurring prohibitive computational expense.

A notable trend in recent MoE research is to aggressively increase the number of experts for finer-grained specialization. Empirical evidence shows that larger expert pools improve overall capacity and often yield higher accuracy with similar or reduced compute costs~\cite{JMLR:v23:21-0998, lepikhin2020gshardscalinggiantmodels}. For instance, PEER~\cite{He2024MixtureOA} can handle millions of experts, and recent theoretical work~\cite{pmlr-v162-clark22a} confirms that MoE performance scales predictably with the expert count.

Following this trend, a natural question arises: \textbf{can we achieve even better performance by further increasing the number of experts to infinity?} In principle, having more experts should allow for even more specialized representations, potentially boosting generalization across diverse tasks. 

We introduce $\infty$-MoE, which moves from a discrete set of experts to a continuous domain, allowing theoretically unbounded expert capacity. In this framework, each input samples from a continuum of experts, taking the concept of “increasing experts” to the extreme.

Despite the potential for an infinite number of experts, our proposed $\infty$-MoE remains computationally tractable because of its sparse activation of only a small number of sampled experts at any given time. This design preserves the efficiency of sparse routing while offering significantly enhanced model capacity. Through experiments on GPT-2 Small and Medium~\cite{radford2019language}, we observed that the GPT-2 Small-based $\infty$-MoE variant (129 million active parameters, 186 million total) achieves performance comparable to that of a dense GPT-2 Medium model with 350 million parameters. Furthermore, increasing the number of samples during inference yields additional accuracy gains, and reducing it still maintains a 2.5\% accuracy improvement over standard MoE, enabling flexible tradeoffs between speed and accuracy.

\section{Related Work}

MoE was first proposed to split a problem space into multiple specialized expert networks~\cite{article}, and has lately gained popularity for LLMs. 
A central advantage in LLMs is that routing each token to just a few experts can greatly expand parameter capacity without a matching increase in compute~\cite{shazeer2017outrageouslylargeneuralnetworks, lepikhin2020gshardscalinggiantmodels, JMLR:v23:21-0998}. For instance, GShard~\cite{lepikhin2020gshardscalinggiantmodels} and Switch Transformer~\cite{JMLR:v23:21-0998} employ sparse expert activation to train models with hundreds of billions of parameters, though they typically rely on a small expert pool (16 to a few hundred) that restricts specialization.

Recent work addresses this issue by substantially raising the expert count. PEER~\cite{He2024MixtureOA} scales up to a million experts, demonstrating richer specialization by novel routing mechanisms. Theoretically, increasing experts improves performance without linearly increasing compute~\cite{pmlr-v162-clark22a, pmlr-v235-ludziejewski24a}, but router overhead can grow large or over-compressed experts may degrade accuracy~\cite{pmlr-v235-ludziejewski24a}.

\begin{table*}[t!]
\caption{Zero-shot performance on various benchmarks (BoolQ~\cite{clark-etal-2019-boolq}, HellaSwag~\cite{zellers-etal-2019-hellaswag}, WinoGrande~\cite{10.1145/3474381}, ARC-e/c~\cite{boratko-etal-2018-systematic}, OpenBookQA~\cite{banerjee-etal-2019-careful}, and RACE-high ~\cite{lai-etal-2017-race}). ``Active/Total Param'' indicates the approximate number of parameters used during forward vs.\ total parameters.}
\label{tab:performance_results}
\centering
\resizebox{\textwidth}{!}{%
\begin{tabular}{lccccccccc}
\toprule
\multirow{1}{*}{\textbf{Model}} 
& \multirow{1}{*}{\textbf{Active/Total Param}} 
& \textbf{BoolQ(↑)} & \textbf{HellaSwag(↑)} & \textbf{WinoGrande(↑)} & \textbf{ARC-e(↑)} & \textbf{ARC-c(↑)} & \textbf{OBQA(↑)} & \textbf{RACE-high(↑)} & \multirow{1}{*}{\textbf{Avg(↑)}} \\
\midrule
\multicolumn{9}{c}{\textbf{GPT-2 Small}} \\
\midrule
Dense & 124M/124M & 60.1 & 29.2 & 50.8 & 43.1 & \textbf{19.4} & 15.2 & 51.3 & 38.5 \\
Switch Transformer & 124M/181M & 60.1 & 29.2 & 51.2 & 43.1 & 18.0 & 14.4 & 51.3 & 38.2 \\
MoE & 124M/181M & \textbf{60.5} & 29.5 & 51.5 & 44.6 & 18.5 & 15.8 & 51.3 & 38.8 \\
$\infty$-MoE & 129M/186M & 59.6 & \textbf{29.8} & \textbf{54.2} & \textbf{46.0} & 18.9 & \textbf{17.6} & \textbf{52.3} & \textbf{39.8} \\
\midrule
\multicolumn{9}{c}{\textbf{GPT-2 Medium}} \\
\midrule
Dense & 350M/350M & 60.7 & 31.4 & 48.8 & 47.1 & 20.1 & 17.6 & 53.1 & 39.8 \\
Switch Transformer & 350M/556M & 58.4 & 31.5 & 50.0 & 48.0 & 20.0 & 16.2 & 55.2 & 39.9 \\
MoE & 350M/556M & \textbf{59.3} & 32.7 & 50.7 & 48.3 & 20.6 & 17.8 & 52.7 & 40.3 \\
$\infty$-MoE & 362M/568M & 56.6 & \textbf{33.7} & \textbf{51.6} & \textbf{49.7} & \textbf{21.5} & \textbf{18.8} & \textbf{57.0} & \textbf{41.3} \\
\bottomrule
\end{tabular}
}
\end{table*}

\section{Proposed Method}
This section presents our $\infty$-MoE framework. We first introduce a generalized MoE formulation for the standard case, then detail the $\infty$-MoE model, which extends MoE to a continuous expert space.

\subsection{Mixture of Experts}

Let $\mathcal{Z} = \{1, 2, \dots, n\}$ be a discrete index set of $n$ experts. Let $x \in \mathbb{R}^{d_{\mathrm{in}}}$ denote the input.  Each expert is a function:
\[
  f(x, i) : \mathbb{R}^{d_{\mathrm{in}}} \times \mathcal{Z} \to \mathbb{R}^{d_{\mathrm{out}}},
\]
where $i \in \mathcal{Z}$ indexes the expert. A router produces a probability distribution $p(i|x)$ over experts.

The MoE output is the expected expert output:
\begin{equation}
\label{eq:general-moe}
  y \;=\; \sum_{i=1}^{n} p(i \mid x) \, f(x,i)
\end{equation}

\paragraph{Connection to Standard MoE.} Standard MoE can be seen as a special case where the general expert function $f(x,i)$ simply selects the $i$-th expert from a set of $n$ pre-defined expert functions, $\{e_1(x), \dots, e_n(x)\}$;  that is, $f(x, i) = e_i(x)$.  The router typically uses a softmax function to compute the probability of selecting expert $i$:
\begin{equation}
p(i|x) = \text{softmax}(TopK(g(x)))_i
\end{equation}
where $g(x) \in \mathbb{R}^n$ is a vector of scores produced by the router network.  With a top-$k$ operation selecting a subset $K$ of experts, the final output is
\begin{equation}
    y =  \sum_{i \in K} p(i|x) \, e_i(x).
\end{equation}
This result clearly demonstrates that the standard MoE is a special case of this discrete formulation.

\subsection{\(\infty\)-MoE: Infinite Experts}
\label{sec:inf-moe}

$\infty$-MoE extends the discrete MoE to a continuous, potentially uncountably infinite, expert space $\mathcal{Z} \subseteq \mathbb{R}^{d_z}$. The router defines a probability density $p(z|x)$ over $\mathcal{Z}$.  The model output is
\begin{equation}
    y = \int_{\mathcal{Z}} p(z \mid x) \, f(x,z)\, dz
\end{equation}
We approximate this integral by Monte Carlo sampling: We sample $z \sim p(z|x)$ and use $f(x,z)$ as an estimator of $y$.  

\paragraph{Router Design.}
\label{sec:router}
We use a Gaussian density for the router:
\begin{equation}
\label{eq:cont_router}
    p(z \mid x) = \mathcal{N}(z \mid \mu(x), \Sigma(x)),
\end{equation}
where a small neural network predicts $\mu(x)$ and $\Sigma(x)$ (i.e., all off-diagonal entries are zero) from $x$. During training, we sample \(z^{(k)} \sim p(z \mid x)\) \(K\) times (\(k=1, \dots, K\)), allowing the router to learn to allocate probability mass to appropriate regions of \(\mathcal{Z}\).

\paragraph{Expert Design.}
\label{sec:expert-function}
We treat \(z\) as a continuous expert index sampled from the router. 
Intuitively, each distinct value of $z$ corresponds to a different expert in an infinite expert space. 
Our feed-forward network (FFN) is then modulated by a mask that ``turns off'' certain neurons in the intermediate layer, allowing the model to dynamically select which subset of parameters is active.

Formally, let $W_z \in \mathbb{R}^{d_{\mathrm{ff}} \times d_{\mathrm{z}}}$. 
Given $z$ sampled from \autoref{eq:cont_router}, we apply a \(\text{top-}N\%\) operator on intermediate neurons $\hat{m}_i=W_z\,z$, which keeps only the largest \(N\%\) of nodes and sets the rest to 0:
\begin{align}
  \mathrm{mask}(z)_i = 
  \begin{cases}
    \hat{m}_i & \text{if } \hat{m}_i \text{ is top } N\%\text{ values},\\
    0 & \text{otherwise}.
  \end{cases}
\end{align}
Because the retained entries preserve their original values, the resulting mask is partially ``soft'' for the selected positions, while all other positions become strictly zero.
Given this mask, the expert output \(f(x,z)\) is computed as
\begin{align}
  f(x, z) &= W_2\Bigl(\mathrm{Act}(W_1 x) \,\odot\, \mathrm{mask}(z)\Bigr),
\end{align}
where \(\mathrm{Act}(\cdot)\) is a nonlinear activation, \(\odot\) is element-wise multiplication, and \(W_1 \in \mathbb{R}^{d_{\mathrm{ff}} \times d_{\mathrm{in}}}, W_2 \in \mathbb{R}^{d_{\mathrm{out}} \times d_{\mathrm{ff}}}\) are learnable weight matrices. 
Through this mechanism, each sampled $z$ effectively activates a distinct subset of the FFN's neurons, mirroring the sparsity in conventional MoE models but generalized to infinite experts.

\begin{figure*}[htbp]
    \centering
    \includegraphics[width=\linewidth]{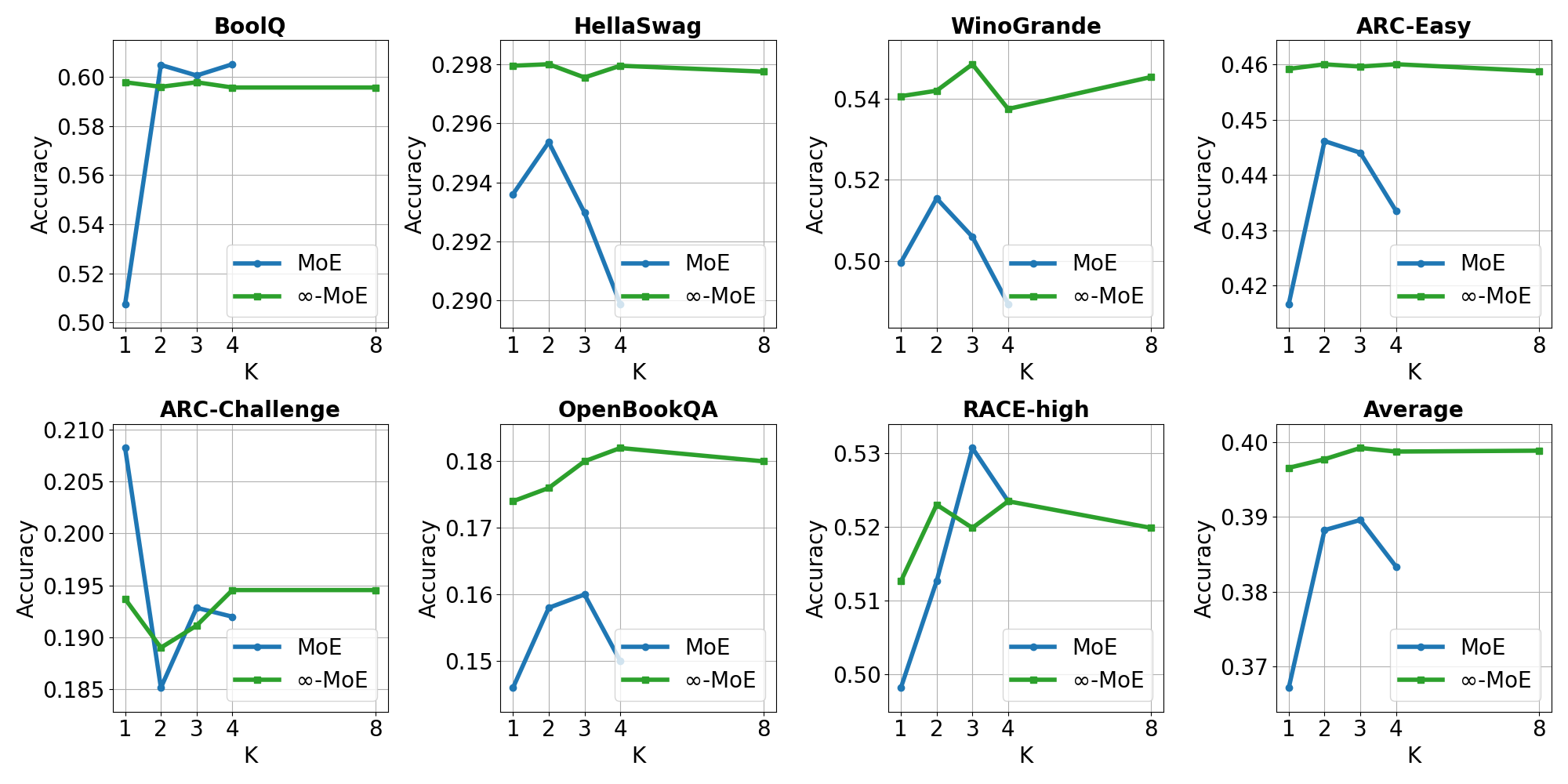}
    \caption{Comparison of MoE and \(\infty\)-MoE models on several tasks while varying the number of experts \(K \in \{1,2,3,4,8\}\). For GPT-2 Small, \(K=2\) yields 124 million active parameters.\(\infty\)-MoE consistently achieves strong accuracy across a wide range of \(K\), even with fewer experts.}
    \label{fig:ablation}
\end{figure*}

\section{Experiments}
We evaluated the effectiveness of $\infty$-MoE using GPT-2 Small ($\sim$124 million parameters) and GPT-2 Medium ($\sim$350 million parameters) on a broad range of natural-language understanding tasks.

\subsection{Setup}
\paragraph{Data.}
We pre-trained our models on a large-scale web corpus called FineWeb~\cite{penedo2024the}, from which we extracted 10~billion tokens. For fine-tuning or direct evaluation, we used the zero-shot setting on standard NLP benchmarks.
\paragraph{Compared Methods.} 

We compared four architectures: \\
\noindent
{\bf \sbt \hspace{1pt} Dense (FFN):} A standard transformer with a single FFN layer shared by all inputs. \newline
{\bf \sbt \hspace{2pt} Switch Transformer (Top-1):} Routes each token to exactly one expert. \newline
{\bf \sbt \hspace{2pt} MoE (Top-2):} A classic sparse MoE setting that activates the top-2 experts for each token. In this configuration, the total number of experts is fixed at 4.  \newline
{\bf \sbt \hspace{2pt} $\infty$-MoE:} Our proposed method with an infinite expert space. During both training and testing, two samples are drawn (i.e., $K=2$); with one sample, only 25\% of the overall expert space is active. \newline

\subsection{Main Results}
Table~\ref{tab:performance_results} presents the zero-shot performance of each model on GPT-2 Small and GPT-2 Medium. Across all tasks, $\infty$-MoE consistently outperformed the Dense baseline, Switch Transformer, and standard MoE. Notably, for GPT-2 Small, $\infty$-MoE achieved the highest average score of 39.8 versus 38.5 (Dense), 38.2 (Switch), and 38.8 (MoE). We observed similar improvements with the GPT-2 Medium variant, where $\infty$-MoE again attained the best average accuracy (41.3).

\section{Ablations}

\subsection{Scaling with sampling ($K$)}
Figure~\ref{fig:ablation} compares $\infty$-MoE with standard MoE across multiple tasks by varying $K$. In the conventional setup, increasing $K$ can improve accuracy but may also introduce instability at high values. By contrast, $\infty$-MoE scales more smoothly with $K$, yielding robust gains and maintaining strong performance even at lower $K$ (achieving a 2.5\% improvement over standard MoE). Moreover, treating experts as a continuous space enables flexible inference, allowing users to adjust $K$ according to hardware constraints or latency requirements.

These results demonstrate that $\infty$-MoE combines the expressiveness of an unbounded expert ensemble with the efficiency of sparse MoE, making it well suited to a variety of runtime conditions.

\subsection{Scaling with Dataset Size}
\label{sec:scaling_size}
\begin{figure}[t!]
    \centering
    \includegraphics[width=0.9\linewidth]{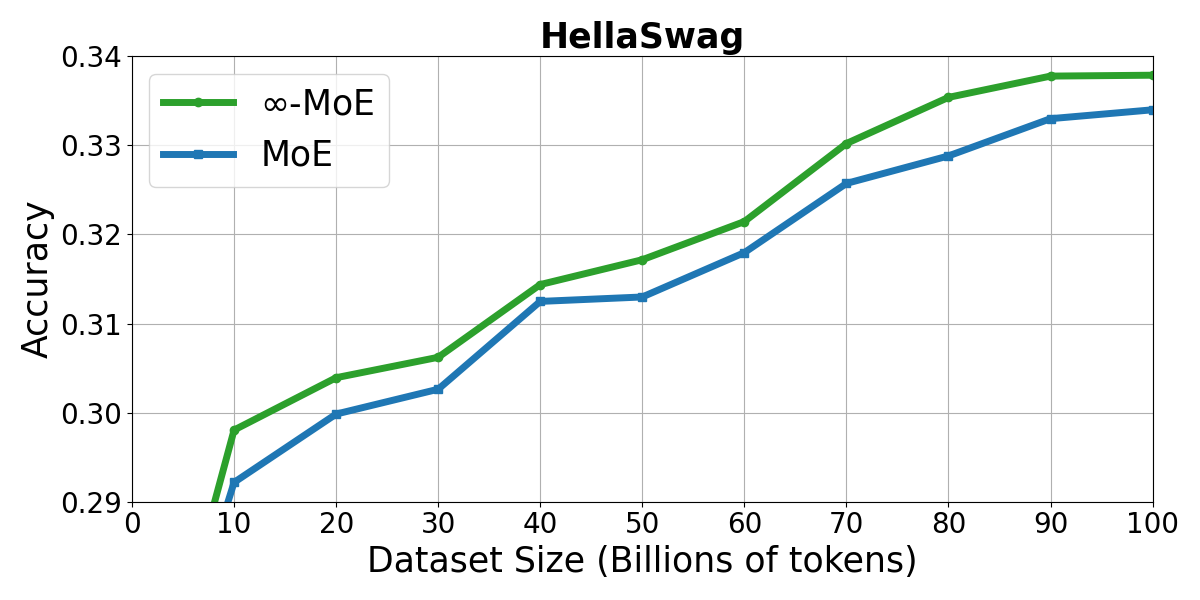}
    \caption{Accuracy on HellaSwag as a function of training data size (in billions of tokens).  $\infty$-MoE is compared against a MoE baseline (GPT-2 Small backbone).}
    \label{fig:data_scale}
\end{figure}

To evaluate the effectiveness of our proposed $\infty$-MoE method under increasing dataset sizes, we conducted experiments using a GPT-2 Small architecture as the base model.  We measured the accuracy on the HellaSwag dataset, progressively increasing the training data size in increments of 10 billion tokens up to 100 billion.  The results were plotted as shown in Figure~\ref{fig:data_scale}.

\subsection{Dimension of Continuous Index $z$}
We varied the dimensionality of the continuous expert index $z$ and report the perplexity in Table~\ref{tab:ablation_z}.
Increasing the dimension improved performance up to $d_z=128$, but larger dimensions provided no further gain.
This result suggests that a moderate index dimension is sufficient to represent diverse experts.

\begin{table}[H]
    \centering
    \caption{Ablation on the dimension of the continuous index $z$. Lower is better.}
    \label{tab:ablation_z}
    \vspace{0.2cm}
    \begin{tabular}{cc}
        \toprule
        \textbf{$d_z$} & \textbf{PPL} \\
        \midrule
        32 & 3.244 \\
        64 & 3.220 \\
        128 & \textbf{3.213} \\
        256 & 3.219 \\
        \bottomrule
    \end{tabular}
\end{table}

\subsection{Routing Stability}
We measured routing stability using the normalized routing entropy.
For each mini-batch, we sampled $z_t^{(k)} \sim p(z \mid x_t)$ ($k=1,\dots,K$) and obtained the active neuron set
$S(z_t^{(k)}) \subseteq \{1,\dots,d_{\mathrm{ff}}\}$ from the top-$N\%$ mask.
We defined the empirical selection distribution over FFN neurons as
\begin{equation}
q_j=
\frac{\sum_{t,k}\mathbf{1}[j \in S(z_t^{(k)})]}
{\sum_{t,k}|S(z_t^{(k)})|}
\quad (j=1,\dots,d_{\mathrm{ff}}),
\end{equation}
and computed
\begin{equation}
\widetilde{H}(q)=\frac{-\sum_{j=1}^{d_{\mathrm{ff}}} q_j \log q_j}{\log d_{\mathrm{ff}}}\in[0,1].
\end{equation}
Higher $\widetilde{H}(q)$ indicates more uniform utilization.
As shown in Table~\ref{tab:entropy}, $\widetilde{H}(q)$ quickly stabilized at a high value during training.

\begin{table}[H]
    \centering
    \caption{Normalized routing entropy $\widetilde{H}(q)$ during training.}
    \label{tab:entropy}
    \vspace{0.2cm}
    \begin{tabular}{cc}
        \toprule
        \textbf{Training Step} & \textbf{$\widetilde{H}(q)$} \\
        \midrule
        5k & 0.966 \\
        10k & 0.902 \\
        15k & 0.893 \\
        20k & 0.893 \\
        \bottomrule
    \end{tabular}
\end{table}

\subsection{Latency Profiling: Custom Kernel}
\begin{figure}[t]
    \centering
    \includegraphics[width=0.9\linewidth]{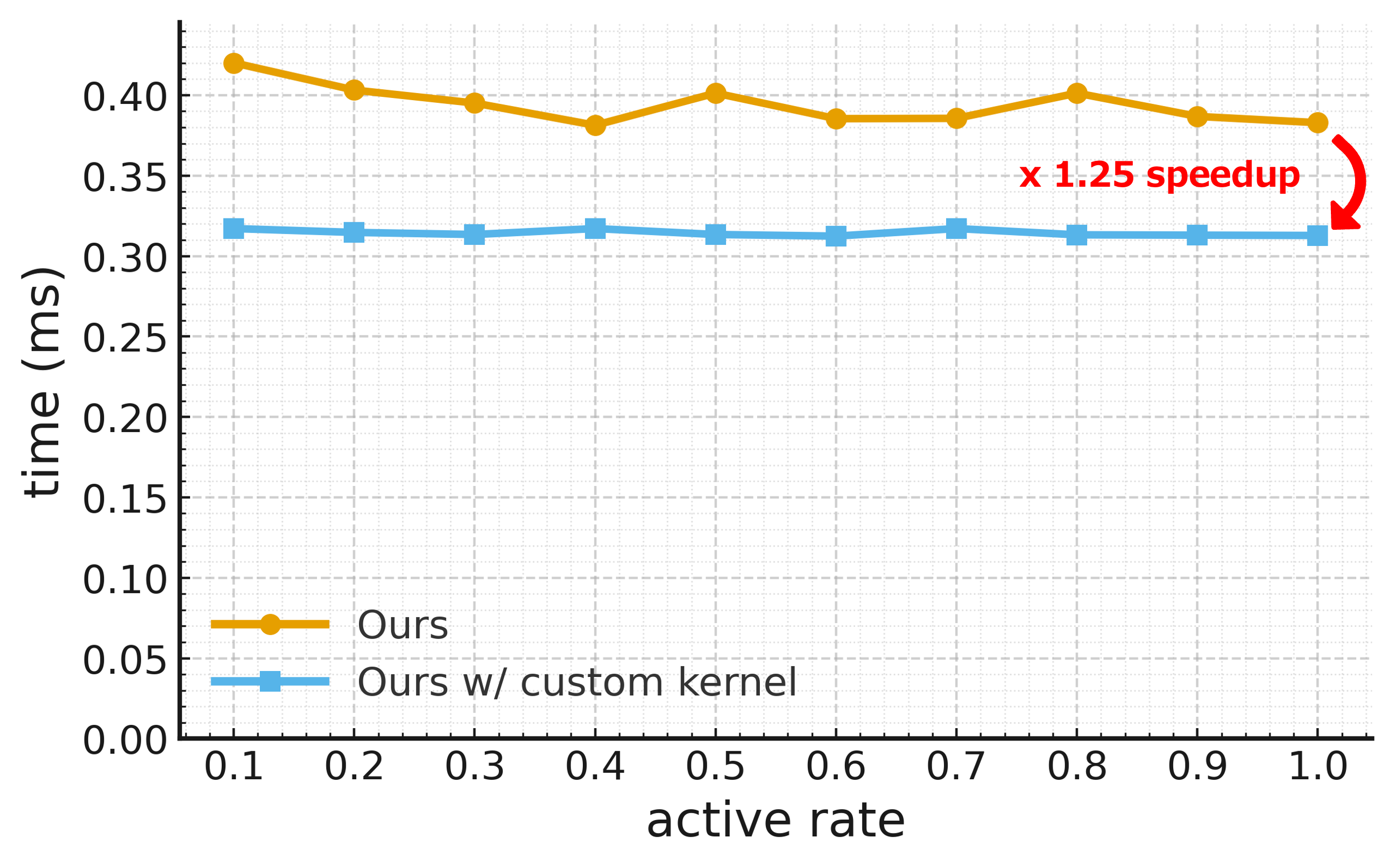}
    \caption{FFN forward latency vs.\ active rate. The custom kernel achieved 1.25-fold speedup.}
    \label{fig:speedtest}
\end{figure}

We measured end-to-end FFN forward latency (ms) while varying the active rate (percentage of FFN hidden units kept by the mask). All runs used the same batch size and sequence length on a single GPU after warmup. We compared a PyTorch reference (Ours) with a fused CUDA implementation that applied the mask and reduction in one pass (Ours w/ custom kernel). Results (Fig.~\ref{fig:speedtest}) show consistently lower latency for the custom kernel and approximately 1.25-fold speedup; latency remained nearly flat across active rates for the custom kernel.

\section{Conclusion}

We present $\infty$-MoE, which extends MoE from a finite set of experts to a continuous (effectively infinite) expert space. It activates only a few sampled experts per token, preserving MoE-like efficiency while improving accuracy. On GPT-2 Small and Medium, $\infty$-MoE outperformed Switch and standard MoE, and tuning the number of samples $K$ at inference provides a clear speed--accuracy tradeoff.

\section*{Limitations}
Although \(\infty\)-MoE offers a promising framework for extending MoE models to an infinite expert space, several open challenges remain:

\begin{enumerate}
\item \textbf{Scaling Beyond GPT-2 Medium.}\\
Although our experiments focused on GPT-2 Small and Medium, the behavior of \(\infty\)-MoE when scaling to larger models (e.g., GPT-3 and beyond) is not yet fully understood.
In particular, it is unclear how performance and efficiency will change when:
\begin{itemize}
    \item Increasing the total number of parameters while keeping the active (per-token) parameter count fixed.
    \item Scaling both active and total parameters in tandem.
\end{itemize}
These scenarios raise questions about potential bottlenecks and tradeoffs in both training and inference at extreme scales.

\item \textbf{Router Distributions.}\\
Our current implementation employs a unimodal Gaussian router for simplicity.
However, richer distributions, such as mixtures of Gaussians or nonparametric density estimators, could offer more expressive expert allocations, especially in high-dimensional expert spaces.
Although this approach may improve coverage of diverse input patterns, designing efficient sampling and sparse-inference mechanisms becomes more complex, and variance reduction in training remains an open challenge.

\item \textbf{Applicability to Other Domains.}\\
Although our study highlights \(\infty\)-MoE's utility in language modeling, it remains unclear how readily this framework generalizes to other domains such as vision (e.g., ViT) or multimodal vision-language models (VLMs).
Practical concerns include adapting continuous expert indices to handle different input modalities, ensuring sparse and efficient routing for high-resolution data, and maintaining competitive accuracy in tasks beyond NLP.

\item \textbf{Training FLOPs and scalability.}
From a compute perspective, $\infty$-MoE preserves the MoE property that per-token FFN FLOPs scale with the activated fraction of parameters.
Theoretically, with $K$ samples and an FFN active ratio $r$ (top-$N\%$ masking), the dominant FFN compute is approximately $\mathcal{O}(K \cdot r)$ of a dense FFN, and the routing overhead is small.
However, lower FLOPs do not always translate to faster training in practice.
Even with our fused CUDA kernel (Figure~\ref{fig:speedtest}), end-to-end throughput can be slower than standard MoE systems because the mask depends on the input token.
Since the set of active hidden units changes across tokens and samples, it is difficult to batch computation into fixed-shape GEMMs(general matrix multiplications, i.e., standard GPU matrix-multiply kernels) and to reuse a single highly optimized kernel.
As a result, naive implementations often rely on extra indexing and accumulation steps, which makes kernel fusion and vectorized execution harder and reduces practical hardware utilization.

In our experiments, we prioritized throughput over strictly minimizing FLOPs, and implemented the FFN computation using masked dense matrix multiplications that include the masking operation in the matmul pipeline.
This design allows using fixed-shape GEMMs with shared weights across samples, enabling highly optimized kernels (e.g., Tensor Core) and improving wall-clock efficiency, at the cost of performing additional arithmetic compared to ideal sparse execution.
Closing the remaining gap between theoretical compute savings and practical speed likely requires further system-level optimizations, as well as better compiler/runtime or hardware support for dynamic sparsity.
\end{enumerate}

\bibliography{latex/custom}

\appendix

\section{Hyperparameter}
Details are provided in Table~\ref{tab:model_training_hyperparameters}.

\begin{table*}[h]
\centering
\caption{Model and training hyperparameters used in the experiments.}
\label{tab:model_training_hyperparameters}
\begin{tabular}{lcccccc}
\toprule
\textbf{Parameter} & \textbf{GPT2-small} & \textbf{GPT2-medium} \\
\midrule
\multicolumn{3}{l}{\textbf{Model Hyperparameters}} \\
\midrule
Block size & 1024 & 1024 \\
Vocab size & 50257 & 50257 \\
Layers & 12 & 24 \\
Heads & 12 & 16 \\
Embedding dim & 768 & 1024 \\
Hidden dim & 3072 & 4096 \\
Gate dim(z dim) & 256 & 256 \\

\midrule
\multicolumn{3}{l}{\textbf{Training Hyperparameters}} \\
\midrule
Total batch size & \multicolumn{2}{c}{524288} \\
Gradient accumulation steps & \multicolumn{2}{c}{1} \\
Optimizer & \multicolumn{2}{c}{\texttt{adamw}} \\
Learning rate & \multicolumn{2}{c}{0.0006} \\
Weight decay & \multicolumn{2}{c}{0.1} \\
Warmup ratio & \multicolumn{2}{c}{0.03} \\
Warmup iterations & \multicolumn{2}{c}{700} \\
Data type & \multicolumn{2}{c}{\texttt{bfloat16}} \\
ZeRO stage & \multicolumn{2}{c}{1} \\
\bottomrule
\end{tabular}
\end{table*}

\section{Total Computation for Experiments}
We executed the experiments mainly by running the training for each model using eight nodes, each equipped with eight NVIDIA H200 (141GB) GPUs.

\section{License}
\subsection{Model}
\begin{itemize}
    \item GPT-2 small/medium: Modified MIT License 
\end{itemize}

\subsection{Dataset}
\begin{itemize}
    \item FineWeb: Open Data Commons Attribution License (ODC-By) v1.0 
\end{itemize}

\section{Additional Experimental Results}

\subsection{Scaling with dataset size}
\label{app:detail_data}
To complement \autoref{sec:scaling_size}, Table~\ref{tab:full_benchmark} reports zero-shot accuracy on the full benchmark suite for GPT-2 Small-based models trained on 100B tokens.
At this larger data scale, $\infty$-MoE achieves a higher average score (0.412 vs.\ 0.407) and improves 5 out of 7 tasks.
These results indicate that the gains from continuous expert indexing persist beyond the 10B-token setting used in the main experiments.

\begin{table*}[h!]
    \centering
    \caption{Comparison of MoE and $\infty$-MoE (Ours) on the full benchmark suite. The models are GPT-2 Small based and trained on 100B tokens.}
    \label{tab:full_benchmark}
    \vspace{0.2cm}
    \resizebox{\textwidth}{!}{%
    \begin{tabular}{lcccccccc}
        \toprule
        \textbf{Model} & \textbf{BoolQ} & \textbf{HellaSwag} & \textbf{WinoGrande} & \textbf{ARC-e} & \textbf{ARC-c} & \textbf{OBQA} & \textbf{RACE-h} & \textbf{Avg} \\
        \midrule
        MoE & 0.592 & 0.334 & \textbf{0.511} & 0.485 & 0.198 & \textbf{0.190} & 0.534 & 0.407 \\
        Ours ($\infty$-MoE) & \textbf{0.605} & \textbf{0.335} & 0.510 & \textbf{0.490} & \textbf{0.206} & 0.170 & \textbf{0.567} & \textbf{0.412} \\
        \bottomrule
    \end{tabular}%
    }
\end{table*}

\subsection{Comparison with More Experts}
To test whether increasing the number of discrete experts is sufficient, we compare $\infty$-MoE with a 16-expert MoE baseline (top-2) trained on 10B tokens.
We keep the active parameter budget the same by using top-2 activation in both models.
As shown in Table~\ref{tab:16_experts_ppl}, $\infty$-MoE achieves lower perplexity, suggesting better parameter efficiency than simply adding more discrete experts.

\begin{table}[H]
    \centering
    \caption{Perplexity comparison with a 16-expert MoE baseline under a matched active-parameter budget (top-2, trained on 10B tokens). Lower is better.}
    \label{tab:16_experts_ppl}
    \vspace{0.2cm}
    \begin{tabular}{lc}
        \toprule
        \textbf{Model} & \textbf{PPL} \\
        \midrule
        MoE (Top-2 over 16 experts) & 3.294 \\
        Ours (Top-2 over infinite experts) & \textbf{3.238} \\
        \bottomrule
    \end{tabular}
\end{table}

\section{Use of Large Language Models}
We employed a large language model (ChatGPT; “GPT-5 Thinking”) only for English‐language polishing and light copy-editing. 
The model was not used to generate ideas and experimental design. 
All technical content and claims were authored and verified by the human authors. 
No non-public data, confidential information, or personally identifiable information were provided to the model. 


\end{document}